# Inertial Sensor-Based Humanoid Joint State Estimation

Nicholas Rotella[1], Sean Mason[1], Stefan Schaal[1,2] and Ludovic Righetti[2]

*Abstract*— This work presents methods for the determination of a humanoid robot's joint velocities and accelerations directly from link-mounted Inertial Measurement Units (IMUs) each containing a three-axis gyroscope and a three-axis accelerometer. No information about the global pose of the floating base or its links is required and precise knowledge of the link IMU poses is not necessary due to presented calibration routines. Additionally, a filter is introduced to fuse gyroscope angular velocities with joint position measurements and compensate the computed joint velocities for time-varying gyroscope biases. The resulting joint velocities are subject to less noise and delay than filtered velocities computed from numerical differentiation of joint potentiometer signals, leading to superior performance in joint feedback control as demonstrated in experiments performed on a SARCOS hydraulic humanoid.

## I. INTRODUCTION

Feedback control for robots relies on accurate estimates of the joint state. Traditionally, the position of each joint is measured using an angular sensor from which joint velocity and acceleration are computed via numerical differentiation. This produces quantities subject to considerable noise, requiring low-pass filtering for use in control. However, filtering introduces a potentially-destabilizing time delay, preventing the use of feedback gains high enough to achieve satisfactory stiffness and damping. Rather than differentiating joint positions, we develop methods for computing joint velocity and acceleration from sensors which measure quantities of the same order. These estimates - derived from gyroscopes and accelerometers - can be used for control directly or through fusion with position sensing in an optimal filtering framework.

While compact and affordable inertial sensors are fairly new, various types of accelerometers have been used in sensor fusion for decades. Schuler et al. [1] were among the first to compute rigid body angular velocity from accelerometers. Padgaonkar et al. [2] introduced methods for computing angular acceleration, with Liu [3] demonstrating that nine axes are needed for stable solutions. Zappa et al. [4] proved that 12 accelerometer axes are actually required to avoid angular velocity singularities.

This research was supported in part by National Science Foundation grants IIS-1205249, IIS-1017134, CNS-0960061, EECS-0926052, the DARPA program on Autonomous Robotic Manipulation, the Office of Naval Research, the Okawa Foundation, the Max-Planck-Society and the European Research Council under the European Union's Horizon 2020 research and innovation programme (grant agreement No 637935). Any opinions, findings, and conclusions or recommendations expressed in this material are those of the author(s) and do not necessarily reflect the views of the funding organizations.

[1] Computational Learning and Motor Control Lab, University of Southern California, Los Angeles, California.
[2] Autonomous Motion Department, Max Planck Institute for Intelligent Systems, Tuebingen, Germany.

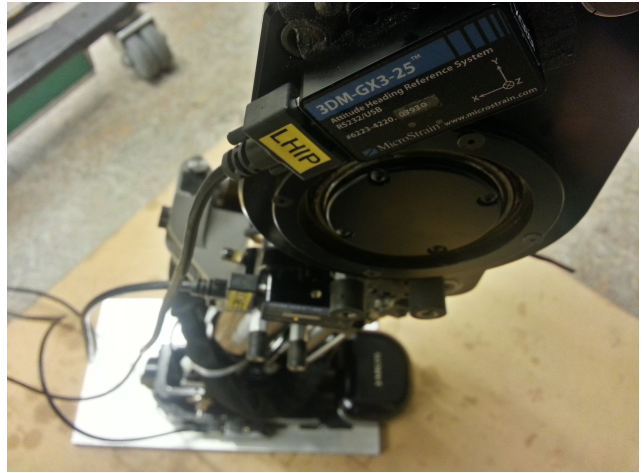

Fig. 1: Inertial Measurement Units (IMUs) attached to the thigh, shank and foot of a SARCOS hydraulic humanoid.

Human joint angle estimation has been researched extensively in the biomechanics community. Williamson et al. [5] attached IMUs to a subject and integrated gyroscope signals to obtain knee joint angles. El-Gohary [6] developed a human arm model used to derive relations between limb IMU measurements and joint state for use in a Kalman Filter. Seel et al. [7] determined joint axes and locations from limb IMUs using knowledge of human kinematic constraints.

Research on robot joint state estimation, however, has been limited. For base state estimation, Lin et al. [8] developed a 12-axis accelerometer suite for their robot, later adding a three-axis gyroscope to avoid singularities. They also proved a three-axis gyroscope plus six accelerometer axes are sufficient to compute angular acceleration if distributed among three distinct locations [9]. For joint angle estimation, Cheng et al. [10] surveyed methods for robots lacking angular sensors. Santaera et al. [11] sensorized manipulator links and used integrated orientation and kinematics to determine joint angles. Vihonen et al. [12] developed an estimator for robots lacking sensors which relied on tilt from accelerometers compensated for inertial effects using gyroscope-based joint velocities, also compensating biases through complementary filtering. In [13], numerical joint accelerations were replaced with estimates from accelerometers and then used to estimate velocities in a complementary filter [14]. Honkakorpi et al. [15] demonstrated that feedback using IMU-based estimates yields similar performance to using encoder measurements. Xinjilefu [16] recently fused predicted joint accelerations, angular velocities from link-mounted gyroscopes and joint states measured by encoders. An orientation calibration routine similar to that presented here is detailed, but accelerometers and joint accelerations are not considered [17]. Unlike previous work, the theory presented here applies to floating base robots with three degree of freedom joints, does not rely on global link orientations, introduces filters for gyroscope bias compensation and considers the use of accelerometers and their role in joint state estimation. Additionally, we present results on feedback control of individual joints which demonstrate the ability to increase both stiffness and damping when using gyroscope-based joint velocities. We begin by considering the theory necessary for computing joint velocities and accelerations from IMU sensor measurements.

## II. SENSOR FRAMEWORK

Assume we have a multibody system composed of $N$ links $L_1, \ldots, L_N$ in series, the first of which is a floating base. These are connected by $N$ joints $J_0, \ldots, J_{N-1}$ where $J_0$ is the floating base orientation in a minimal set of coordinates. We fix an IMU containing a three-axis gyroscope and a three-axis accelerometer to each link at a known position in the link frame and with the same orientation as the link frame (IMU pose calibration is detailed in Sec. V-A and V-B).

Let $\omega_{L_1}^W, \omega_{L_2}^W, \cdots, \omega_{L_N}^W$ denote the angular velocities of each link in the world (global) frame; the gyroscope on $L_i$ thus measures

$$\bar{\omega}_{L_i} = R_W^{L_i}(\omega_{L_i}^W + b_{\omega, L_i} + w_{\omega, L_i}) \quad (1)$$

where $R_W^{L_i}$ rotates a quantity in world frame into the local frame of $L_i$ and $b_{\omega, L_i}$ and $w_{\omega, L_i}$ denote time-varying bias and thermal noise vectors, respectively. Similarly, $a_{L_1}^W, a_{L_2}^W, \cdots, a_{L_N}^W$ denote the true accelerations of the IMU locations on each link in world frame so that the accelerometer on $L_i$ measures

$$\bar{a}_{L_i} = R_W^{L_i}(a_{L_i}^W + b_{a, L_i} + w_{a, L_i} + g) \quad (2)$$

where $g = [0, 0, -9.81]$ is the gravity vector and $b_{a, L_i}$ and $w_{a, L_i}$ again denote bias and noise vectors, respectively. For the time being, we will not consider the effects of noise sources; these will be addressed in Sec. VI.

## III. JOINT VELOCITY COMPUTATION

Assuming in the most general case that every joint has three Degrees of Freedom (DoFs), the vector $\omega_{i-1,i}^i$ measuring the angular velocity of $L_i$ relative to that of $L_{i-1}$ in the $L_i$ frame corresponds to the velocity $\dot{\theta}_{i-1} \in R^3$ of joint $J_{i-1}$. From this point on, we will drop the use of $L$ and $J$ in subscripts for the sake of brevity. For the floating base ($L_1$) gyroscope we thus have

$$\bar{\omega}_1 = R_W^1 \omega_1^W = \dot{\theta}_0$$

where $\dot{\theta}_0$ is the base angular velocity in the base link frame. For $L_2$ we have

$$\bar{\omega}_2 = R_W^2 \omega_2^W = R_W^2(\omega_1^W + \omega_{1,2}^W)$$

where we have used the velocity composition rule. This simplifies further to

$$\bar{\omega}_2 = R_1^2 R_W^1 \omega_1^W + R_W^2 \omega_{1,2}^W = R_1^2 \dot{\theta}_0 + \dot{\theta}_1$$

where $R_1^2$ represents a rotation from frame $L_1$ to $L_2$. Solving for the velocity of $J_1$, we have

$$\dot{\theta}_1 = \bar{\omega}_2 - R_1^2 \dot{\theta}_0$$

Similarly, for $L_3$ we have (again using velocity and rotation composition rules)

$$\begin{aligned}\bar{\omega}_3 &= R_W^3 \omega_3^W \\ &= R_W^3(\omega_1^W + \omega_{1,2}^W + \omega_{2,3}^W) \\ &= R_2^3 R_1^2 R_W^1 \omega_1^W + R_2^3 R_W^2 \omega_{1,2}^W + R_W^3 \omega_{2,3}^W \\ &= R_2^3 R_1^2 \dot{\theta}_0 + R_2^3 \dot{\theta}_1 + \dot{\theta}_2\end{aligned}$$

and solving for the velocity of $J_2$ yields

$$\dot{\theta}_2 = \bar{\omega}_3 - R_2^3 R_1^2 \dot{\theta}_0 - R_2^3 \dot{\theta}_1$$

Continuing in this manner, we form the linear system

$$\begin{bmatrix} I & 0 & 0 & \cdots & 0 \\ R_1^2 & I & 0 & & \vdots \\ R_1^3 & R_2^3 & \ddots & \ddots & \vdots \\ \vdots & \vdots & \ddots & I & 0 \\ R_1^N & R_2^N & \cdots & R_{N-1}^N & I \end{bmatrix} \begin{bmatrix} \dot{\theta}_0 \\ \dot{\theta}_1 \\ \vdots \\ \dot{\theta}_{N-2} \\ \dot{\theta}_{N-1} \end{bmatrix} = \begin{bmatrix} \bar{\omega}_1 \\ \bar{\omega}_2 \\ \vdots \\ \bar{\omega}_{N-1} \\ \bar{\omega}_N \end{bmatrix} \quad (3)$$

which we can write more compactly as $T\dot{\theta} = \bar{\omega}$ where $\dot{\theta} \in R^{3N}$ and $\bar{\omega} \in R^{3N}$ are the full vectors of joint velocities and link gyroscope measurements, respectively. Note that $T$ is a function of $\theta$ and is composed of relative rotations between links; unlike in other methods, there is no need for global link poses. Additionally, since $T$ is lower-triangular this system always has a unique solution which can be determined efficiently using forward substitution.

### A. Computing Constrained Velocities

The above assumes that every joint has three DoFs; however, it is often the case in for a robot that certain joints have fewer (for example, the knee of a humanoid). In this case, more accurate solutions may result from properly constraining velocities using knowledge of the kinematic structure.

For the $L_i$ gyroscope we again have (neglecting noise)

$$\bar{\omega}_i = R_W^i \omega_i^W = R_W^i J_i^W \dot{q} = J_i^1 \dot{q}$$

where $\dot{q} = [\omega_{base}, \dot{q}_{joints}]$ is the vector of generalized joint velocities and $J_i^1$ is the Jacobian relating the angular velocity of $L_i$ to generalized joint velocities in the $L_i$ frame. Using this relation for each link, we form the system

$$\begin{bmatrix} J_1^1 \\ J_2^2 \\ \vdots \\ J_N^N \end{bmatrix} \dot{q} = \begin{bmatrix} \bar{\omega}_1 \\ \bar{\omega}_2 \\ \vdots \\ \bar{\omega}_N \end{bmatrix} \quad (4)$$

where each Jacobian above is computed using Jacobians relative to the base as $J_i^i = R_1^i J_i^1$. We can write this system as $T_J \dot{q} = \bar{w}$ and solve a least-squares problem for joint velocities. While forming the matrix $T_J$ requires knowledge of the kinematic structure, this method results in velocities which are properly constrained and expressed in the correct frames. We will demonstrate the effect of properly constraining velocities in Sec. VII.

## IV. JOINT ACCELERATION COMPUTATION

We now use the setup detailed [1] in Sec. II to determine the acceleration $\ddot{\theta}_{i-1}$ of $J_{i-1}$. The IMU accelerometer on the preceding link measures

$$\bar{a}_{i-1} = R_W^{i-1}(a_{i-1}^W + g)$$

and the accelerometer on the link immediately following $J_{i-1}$ measures

$$\bar{a}_i = R_W^i(a_i^W + g) = R_W^i(a_{i-1}^W + a_{i-1,i}^W + g)$$

where $a_{i-1,i}^W$ is the acceleration of the $L_i$ IMU with respect to the $L_{i-1}$ IMU and is given by

$$a_{i-1,i}^W = (\omega_{i-1,i}^W \times (\omega_{i-1,i}^W \times r_{i-1,i}^W)) + (\alpha_{i-1,i}^W \times r_{i-1,i}^W)$$

where $r_{i-1,i}^W$, $\omega_{i-1,i}^W$ and $\alpha_{i-1,i}^W$ are the position, angular velocity and angular acceleration of the $L_i$ IMU with respect to the $L_{i-1}$ IMU in world frame. Using the notation $a \times b = a^\times b$ where $a^\times \in R^{3\times 3}$, it follows that

$$\begin{aligned}\bar{a}_i &= R_{i-1}^i \left[R_W^{i-1}(a_{i-1}^W + g)\right] + R_W^i a_{i-1,i}^W \\ &= R_{i-1}^i \bar{a}_{i-1} + R_W^i \left[((\omega_{i-1,i}^W)^\times)^2 r_{i-1,i}^W + (\alpha_{i-1,i}^W)^\times r_{i-1,i}^W\right] \\ &= R_{i-1}^i \bar{a}_{i-1} + ((\omega_{i-1,i}^i)^\times)^2 r_{i-1,i}^i - (r_{i-1,i}^i)^\times \alpha_{i-1,i}^i \\ &= R_{i-1}^i \bar{a}_{i-1} + ((\dot{\theta}_{i-1})^\times)^2 r_{i-1,i}^i - (r_{i-1,i}^i)^\times \ddot{\theta}_{i-1}\end{aligned}$$

where we have used the definition of $\bar{a}_{i-1}$, properties of the cross product and the definitions $\dot{\theta}_{i-1} = \omega_{i-1,i}^i$ and $\ddot{\theta}_{i-1} = \alpha_{i-1,i}^i$. We rearrange the above equation to get

$$(r_{i-1,i}^i)^\times \ddot{\theta}_{i-1} = R_{i-1}^i \bar{a}_{i-1} - \bar{a}_i + ((\dot{\theta}_{i-1})^\times)^2 \quad \forall i = 2,\ldots,N$$

Note that gravity does not appear in the above, making gravity compensation unnecessary. Also, the joint velocities $\dot{\theta}_i$ are determined from the link gyroscopes as in Sec. III. However, we cannot solve this equation for $\ddot{\theta}_{i-1}$ because the skew-symmetric matrix $(r_{i-1,i}^i)^\times \in R^{3\times 3}$ has rank two.[2] We thus add a second IMU to $L_i$ and solve

$$\begin{bmatrix}(r_{i-1,i}^i)^\times \\ (\tilde{r}_{i-1,i}^i)^\times\end{bmatrix} \ddot{\theta}_{i-1} = \begin{bmatrix}R_{i-1}^i \bar{a}_{i-1} - \bar{a}_i + ((\dot{\theta}_{i-1})^\times)^2 \\ R_{i-1}^i \bar{a}_{i-1} - \tilde{\bar{a}}_i + ((\dot{\theta}_{i-1})^\times)^2\end{bmatrix} \quad (5)$$

where $\tilde{r}_{i-1,i}^i$ and $\tilde{\bar{a}}_i$ denote the relative position and the acceleration of the additional IMU. The matrix multiplying $\ddot{\theta}_{i-1}$ now has rank three aside from singular cases (for example when $\tilde{r}_{i-1,i}^i = r_{i-1,i}^i$, implying that the IMU positions should be as distinct as possible). Note that rather than use two three-axis accelerometers, we could have equivalently used three two-axis accelerometers to obtain a full-rank matrix; these results are in agreement with [9].

---

[1] Note that we could again constrain the acceleration using the Jacobian since $\alpha = J\ddot{\theta} + \dot{J}\dot{\theta}$ but we wish to avoid using the Jacobian derivative, which must be computed numerically and is thus typically very noisy.

[2] Real skew-symmetric matrices have purely imaginary eigenvalues, which must come in conjugate pairs; thus, the rank must be even.

## V. IMU POSE CALIBRATION

To obtain accurate estimates of joint velocities and accelerations, the IMUs must be fixed with known poses. As is often the case for humanoids, the base IMU is rigidly fixed with a known pose; we compute orientation and position offsets for each link relative to this IMU using the following principle. When rotated in the air with locked joints, the entire robot becomes a single rigid body subject to the same angular velocity and angular acceleration. This is the basis for the following offline calibration methods which recover the full pose (orientation and position) of each IMU with respect to its local link frame.

### A. Orientation Calibration

When every link has the same angular velocity, we can equate the velocities of $L_i$ and that of the base to obtain

$$\hat{R}_i \bar{\omega}_i = R_1^i \bar{\omega}_1$$

where $\hat{R}_i$ is the desired rotational correction for the IMU on $L_i$. Transposing both sides and stacking $M$ consecutive measurements, we obtain

$$\begin{bmatrix}\{\bar{\omega}_i^T\}_1 \\ \{\bar{\omega}_i^T\}_2 \\ \vdots \\ \{\bar{\omega}_i^T\}_M\end{bmatrix} \hat{R} = \begin{bmatrix}\{\bar{\omega}_1^T (R_1^i)^T\}_1 \\ \{\bar{\omega}_1^T (R_1^i)^T\}_2 \\ \vdots \\ \{\bar{\omega}_1^T (R_1^i)^T\}_M\end{bmatrix}$$

where $\{v\}_m$ denotes the $m^{th}$ observation of quantity $v$. We seek $\hat{R}_i$ as the solution to the problem

$$\hat{R} = \arg\min_X ||AX - B||_F^2$$

subject to $X^T X = I$ where $||A||_F$ denotes the Frobenius matrix norm. This is called the orthogonal Procrustes problem and is solved by computing the SVD of $A^T B = U\Sigma V^T$ and setting $\hat{R} = UV^T$. In order to ensure that the solution is a proper rotation matrix we also require $\det(\hat{R}) = +1$. This is known as the Kabsch algorithm [18] and is achieved by instead setting $\hat{R} = U\hat{\Sigma}V^T$ where $\hat{\Sigma} = \text{diag}(1, 1, \text{sign}(\det(UV^T)))$.

### B. Position Calibration

Assuming every link has the same world frame angular acceleration $\alpha^W$, the linear acceleration of the IMU on $L_i$ relative to that of the base link IMU is

$$a_i^W - a_1^W = \omega^W \times \omega^W \times r_{1,i}^W + \alpha^W \times r_{1,i}^W$$

However, we know that $\bar{a}_i = R_W^i(a_i^W + g)$ and thus $a_i^W = R_i^W \bar{a}_i - g$ so

$$a_i^W - a_1^W = R_i^W \bar{a}_i - g - R_1^W \bar{a}_1 + g = R_1^W(R_i^1 \bar{a}_i - \bar{a}_1)$$

Note that the gravity vector again cancels, making compensation unnecessary. Multiplying both sides by $R_W^1$, we have

$$R_i^1 \bar{a}_i - \bar{a}_1 = \omega^1 \times \omega^1 \times r_{1,i}^1 + \alpha^1 \times r_{1,i}^1$$

which we can rewrite in the form

$$[(\bar{\omega}^\times)^2 + \bar{\alpha}^\times] r_{1,i}^1 = (R_i^1 \bar{a}_i - \bar{a}_1)$$

where $\bar{\omega}$ and $\bar{\alpha}$ denote the angular velocity and acceleration measured by the base link IMU. Again, stacking $M$ consecutive measurements results in the linear system

$$\begin{bmatrix} \{(\bar{\omega}^\times)^2 + \bar{\alpha}^\times\}_1 \\ \{(\bar{\omega}^\times)^2 + \bar{\alpha}^\times\}_2 \\ \vdots \\ \{(\bar{\omega}^\times)^2 + \bar{\alpha}^\times\}_M \end{bmatrix} r_{1,i}^1 = \begin{bmatrix} \{R_i^1 \bar{a}_i - \bar{a}_1\}_1 \\ \{R_i^1 \bar{a}_i - \bar{a}_1\}_2 \\ \vdots \\ \{R_i^1 \bar{a}_i - \bar{a}_1\}_M \end{bmatrix}$$

Note that we compute $\bar{\alpha}$ numerically from base gyroscope measurements. Since this is an offline calibration routine, we apply a zero-delay filter to the measurements. Since the matrix multiplying IMU position is the same for every link, we compute its SVD once and find the least-squares solution for each IMU. After solving for the position $r_{1,i}^1$ of the $i^{th}$ IMU in the base frame, we compute its local link position from kinematics.

## VI. JOINT STATE FILTERS

In this section, we introduce two Kalman Filters for the joint state which fuse joint velocities and accelerations computed from inertial sensors with joint position sensor measurements. This is advantageous over directly using the computed joint state because it ensures consistency between the joint position and its derivatives. In theory, performing filtering using an accurate process model also creates less delay than simply low-pass filtering computed quantities.

*Joint Position and Gyroscope Bias Filter*

Assume now that each gyroscope is afflicted by a time-varying bias and thermal noise as in the model given by Eq. (1). From Eq. (4) we thus have

$$T_J(\theta)\dot{\theta} = \bar{\omega} - b - w$$

where $b, w \in R^{3N}$ are the bias and noise vectors for all gyros, respectively. Combining the joint positions and gyroscope biases into the state vector $x = [\theta^T, b^T]^T \in 6N$, we can write the state dynamics as

$$\dot{x} = f(x, \bar{\omega}, w, w_b)$$

or more specifically as

$$\dot{\theta} = -T_J(\theta)^{-1}b + T_J(\theta)^{-1}(\bar{\omega} - w)$$
$$\dot{b} = w_b$$

where we have modeled the bias dynamics as Brownian motion (the integral of white noise process $w_b$). Since we have angular position sensors, we can write a measurement of the state as

$$y = \begin{bmatrix} I & 0 \end{bmatrix} \begin{bmatrix} \theta \\ b \end{bmatrix} + v$$

where $v \in R^{3N}$ is the measurement noise afflicting the joint position sensor measurements. Since the process model is nonlinear, we choose an Extended Kalman Filter (EKF) for implementation.

*Acceleration-Based Joint Velocity Filter*

In order to filter joint velocities, we need joint accelerations $\ddot{\theta}$ for use in the process model. Defining the state vector $x = [\theta^T, \dot{\theta}^T]^T = [x_1^T, x_2^T]^T \in 6N$ we have the trivial dynamics

$$\dot{x}_1 = x_2$$
$$\dot{x}_2 = \ddot{\theta}$$

The joint acceleration vector $\ddot{\theta}$ can be specified in any the following ways, depending on sensor availability:

$$\ddot{\theta} = \begin{cases} M(\theta)^{-1}[\bar{\tau} + J(\theta)_c^T \bar{F} - n(\theta, \dot{\theta})] \\ f(\theta, \dot{\theta}, \bar{a}) \\ \ddot{\theta}_{des} \end{cases}$$

The first method uses measured joint torques $\bar{\tau}$ and endeffector wrenches $\bar{F}$ along with the robot's dynamic model to compute the current joint accelerations. The second method solves for accelerations using link accelerometers with Eq. (5). Alternatively, we can simply use the desired or predicted joint accelerations $\ddot{\theta}_{des}$ from the current controller. Note that use of the first two methods results in a nonlinear process model; in this work, we consider only using desired accelerations for simplicity.

We measure the joint positions from angular sensors and the joint velocities from the link IMUs as in Sec. III-A, leading to the measurement model $y = x + v$ where $v \in R^{6N}$ is again the measurement noise vector.

## VII. EXPERIMENTS AND RESULTS

The platform used for experiments is the lower body of a SARCOS hydraulic humanoid robot having a total of 14 DoFs (seven per leg). An IMU containing a three-axis gyroscope and three-axis accelerometer was mounted to each link - one on the base and one on the link immediately following the hip, knee and ankle joints. We chose Microstrain 3DM-GX3-25 IMUs for their USB interface, low-noise sensors and maximum streaming rate of 1 kHz. Initial gyroscope biases were removed at startup. The IMUs were fixed using an adhesive; while an effort was made to align the IMU axes with the link frames, using the calibration methods of Sec. V-A and V-B allowed for imprecise sensor placement.

We first evaluated the joint velocity computation detailed in Sec. III during a sine tracking task for every joint in the right leg. Joint velocities were computed from measured gyroscope data in three ways: A) using Eq. (3) to compute orientation-corrected but unconstrained velocities, B) using Eq. (4) to compute constrained but non-calibrated velocities and C) again using Eq. (4) to compute velocities both orientation-corrected and constrained to the kinematics. These were compared against velocities computed from potentiometer measurements and filtered using a second-order Butterworth filter with a cutoff of $25Hz$. Fig. 2 shows the effect of constraining computed joint velocities; the constrained hip velocities more closely match the filtered potentiometer-based velocities. Fig. 3 shows the effect of performing the offline rotation calibration routine on constrained ankle velocities; the benefit of calibration is apparent.

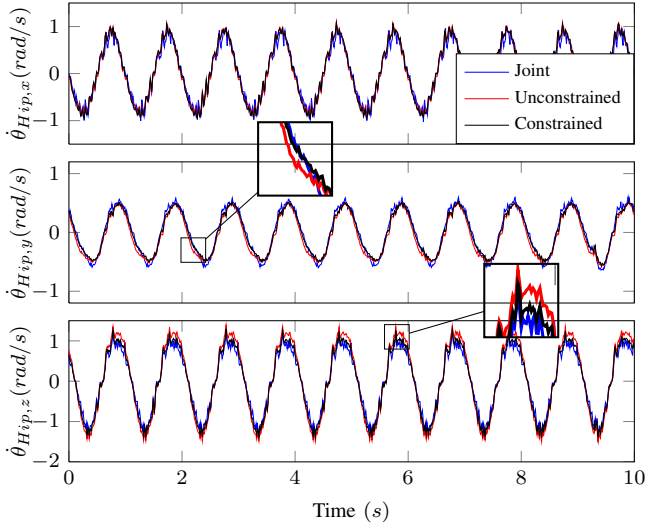

Fig. 2: Comparison between filtered potentiometer-based hip velocities and constrained versus unconstrained hip velocities computed from link gyroscopes.

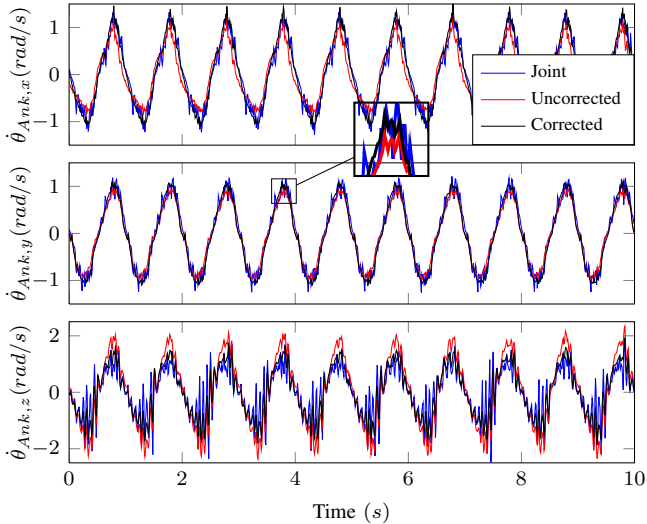

Fig. 3: Comparison between filtered potentiometer-based ankle velocities and rotation-corrected versus uncorrected ankle velocities computed from link gyroscopes.

We next evaluate the IMU position calibration and accelerometer-based joint acceleration computations detailed in Sec. IV. Fig. 4 compares the potentiometer-based hip accelerations (heavily filtered at a cutoff of $5Hz$) with those computed from IMUs using Eq. (5) during a sine task for the hip. We compare the results of computing accelerations using manually-measured IMU positions and automatically-generated positions from the calibration routine of Sec. V-B. Both signals estimate the acceleration well and with much less noise than unfiltered, numerically-computed joint accelerations (shown for reference in Fig. 5). The filtered joint-based accelerations are clearly heavily delayed; this is most evident in the $z$ direction. Because we only have one IMU per link, we compute hip joint accelerations using the base, thigh and shank IMUs with the knee locked. This is not ideal since these IMUs are not truly on the same link; we expect to compute accelerations which are more accurate and less noisy by adding a second IMU to the proper link.

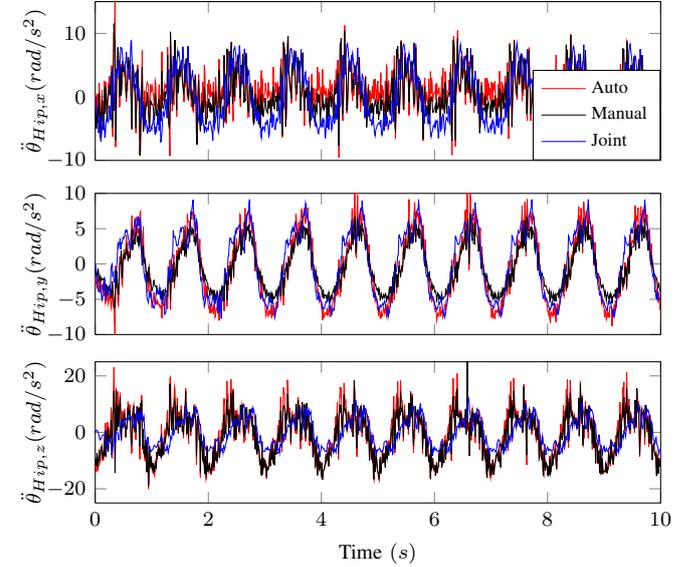

Fig. 4: Filtered potentiometer-based hip joint accelerations versus those computed from inertial sensors with both manually and automatically-generated IMU position information.

It should be noted that obtaining a good position calibration requires sufficient angular motion of the robot, which can be difficult depending on the setup. Additionally, the position calibration and angular acceleration computations rely on accurate kinematics and a good IMU orientation calibration, else the gravity terms will not cancel; these appear in Eq. (5) to create configuration-dependent acceleration biases. Unmodeled accelerometer biases will have the same effect and should be compensated through calibration, however were neglected here.

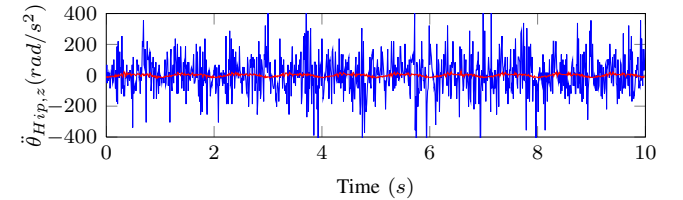

Fig. 5: Raw potentiometer-based hip joint acceleration (blue) versus that computed from inertial sensors (red).

In order to test the gyroscope bias estimator presented in Sec. VI, we performed experiments in the SL simulation environment by simulating gyroscopes subject to the noise sources in Eq. (1). This was done because the IMUs on our robot have a low bias instability; however, this is not the case for inexpensive IMUs which are cost-effective to use in sensorizing every link. We simulate aggressive gyroscope biases which are initially nonzero and drift orders of magnitude faster than those in our IMUs. Fig. 6 shows the results for one of the simulated IMUs. Despite being run during a full-body sine tracking task with simulated joint sensor noise, the filter manages to track the true biases.

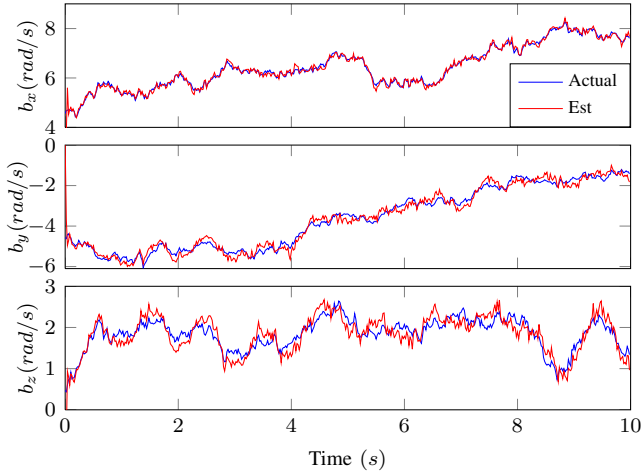

Fig. 6: Simulated gyroscope bias estimation using the joint state filter of Sec. VI.

We also test the joint velocity estimator of Sec. VI which was implemented for all 14 joint DoFs. Fig. 7 compares the joint velocity of one hip DoF for the filtered potentiometer-based velocity, the constrained IMU-based velocity from Eq. (4), and velocities from the estimator of Sec. VI both without and with desired joint accelerations as process model inputs. Both filtered velocities are smoother than the IMU-based velocity, however the estimate from the filter which uses desired acceleration in its process model (in black) has tens of milliseconds less delay than the estimate from the filter having a naive process model (in green below). The desired acceleration-based estimator provides a filtered signal with only a slight delay compared to the IMU-computed velocity. Given the apparent difference between the joint accelerations shown in Fig. 4 and the sinusoids used to generate them, we expect velocity filter performance to improve considerably by using either sensor-based joint accelerations or accelerations computed using the dynamic model. We leave this investigation to future work.

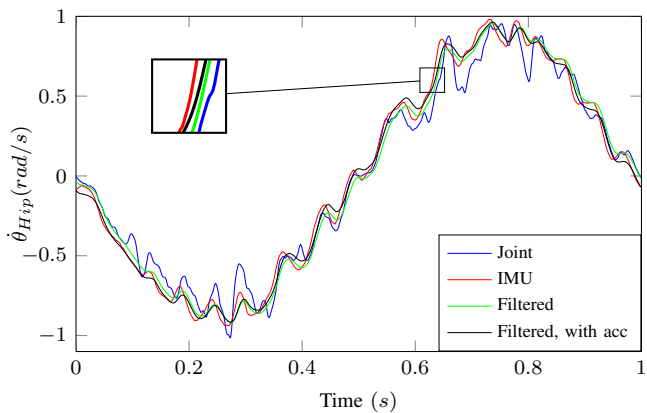

Fig. 7: Hip joint velocity computed from filtered joint sensors, directly from gyroscope velocities, filtered using the estimator of Sec. VI without and with desired accelerations.

Finally, we perform sine tracking tasks for a single joint in isolation (here the right knee) in order to determine whether using the IMU-based joint velocities allows us to increase controller gains and enable better tracking. A joint proportional-derivative (PD) controller was implemented for the knee using potentiometer-based joint position and velocity (filtered at a $25Hz$ cutoff) with the ability to switch to using the raw joint position and gyroscope-based velocity.

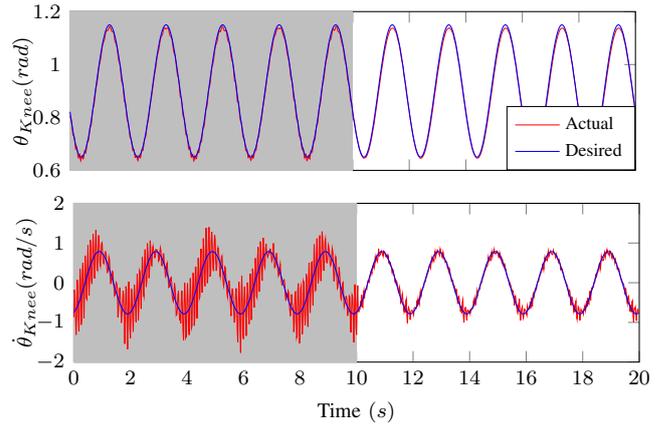

Fig. 8: Knee sine tracking for the gains $P = 1000$ and $D = 12$, switched from potentiometer-based velocities to IMU-based velocities at $t = 10s$.

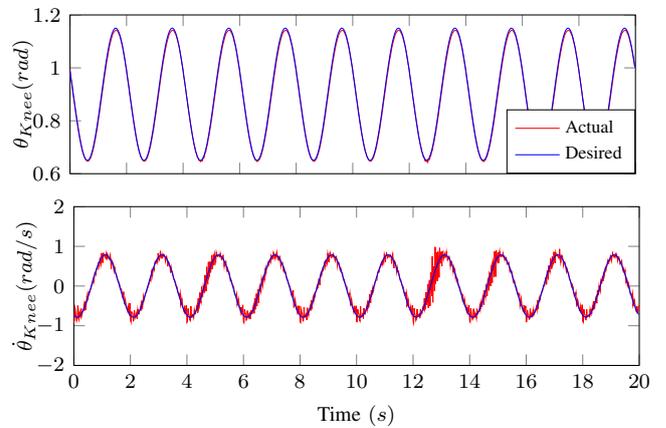

Fig. 9: Knee sine tracking for the gains $P = 1500$ and $D = 12$ for the IMU-based velocities only.

For a $0.5Hz$ sine wave of amplitude $0.25rad$, we were able to increase the position gain to 1000 before the controller using the potentiometer-based velocity went unstable while we could increase the gain to 1600 before the controller using the velocity computed from Eq. (4) showed signs of instability. Fig. 8 shows the tracking using a position gain of 1000 and a velocity gain of 12 for both potentiometer and gyroscope-based joint velocities. RMS tracking error decreases from $0.0103rad$ to $0.0099rad$ in position and from $0.3786rad/s$ to $0.0902rad/s$ in velocity by switching to gyroscope-based velocities. Fig. 9 shows the tracking using a position gain of 1500 and a velocity gain of 12 for the gyroscope-based velocities, demonstrating stable position tracking with an RMS error of $0.0077rad$.

We were also able to independently increase the knee velocity gain from a maximum stable value of 26 using the potentiometer-based velocities to 30 using the gyroscope-based velocities. Fig. 10 shows the tracking for $D = 26$ using each of these velocities. Position tracking is poor using both since the position gain was held at 250, however it is clear from the figure that damping is improved after the switch.

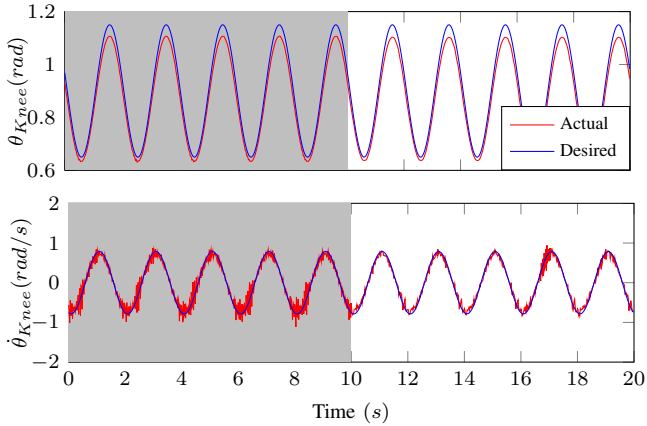

Fig. 10: Knee sine tracking for $P = 250$ and $D = 26$, switched to IMU-based velocities at $t = 10s$.

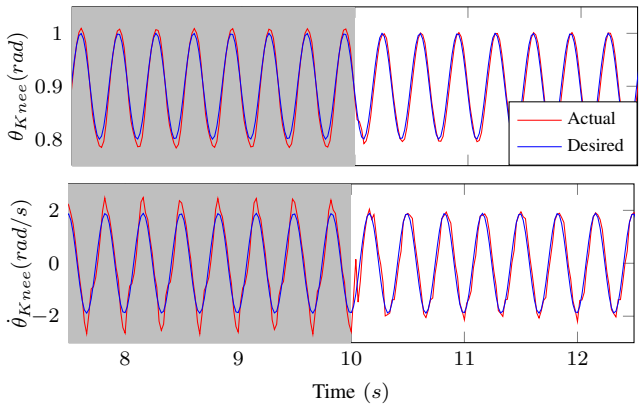

Fig. 11: Knee tracking a 3Hz sine, switched from potentiometer-based velocities with $P = 800$ and $D = 12$ to IMU-based velocities with $P = 1500$ at $t = 10s$.

We finally demonstrate that the above results hold for faster sine tracking by performing a $3Hz$ tracking task using position gains close to the stable limit for each of the potentiometer and gyroscope-based knee velocities. It is evident from Fig. 11 that both position and velocity tracking are improved after switching to the IMU-based velocities due to the ability to use higher gains.

## VIII. CONCLUSIONS

In this paper, we have presented methods for computing joint velocities and accelerations directly from inertial sensor measurements. Offline calibration procedures allow for accurate recovery of the pose of the IMUs attached to each link of the robot. Two filters were proposed in order to fuse measured joint positions with IMU sensor data, allowing for accurate estimation of gyroscopes biases and improvements in the quality of the joint position and velocity estimation. The presented experiments showed that with our method we could improve the quality of the velocity and acceleration estimates, allowing us to significantly increase the stiffness and damping of our joint feedback controllers; this demonstrates the utility of these methods for use in the control of a humanoid robot. Future work will include verification of the proposed approaches using a ground-truth reference such as a VICON system as well as their use in more sophisticated whole-body control loops.

## REFERENCES

[1] A. R. Schuler, A. Grammatikos, and K. A. Fegley, "Measuring rotational motion with linear accelerometers," *Aerospace and Electronic Systems, IEEE Transactions on*, vol. AES-3, no. 3, pp. 465–472, May 1967.

[2] A. Padgaokkar, K. W. Krieger, and A. I. King, "Measurement of angular acceleration of a rigid body using linear accelerometers," *ASME Journal of Applied Mechanics*, vol. 42, no. 3, pp. 552–556, 1975. [Online]. Available: http://dx.doi.org/10.1115/1.3423640

[3] Y. K. Liu, "Discussion: measurement of angular acceleration of a rigid body using linear accelerometers," *ASME Journal of Applied Mechanics*, vol. 43, no. 2, pp. 377–378, 1976. [Online]. Available: http://dx.doi.org/10.1115/1.3423861

[4] B. Zappa, G. Legnani, and A. J. van den Bogert, "On the number and placement of accelerometers for angular velocity and acceleration determination," *Journal of Dynamic Systems, Measurement, and Control*, vol. 123, pp. 552–554, 2001.

[5] R. Williamson and B. Andrews, "Detecting absolute human knee angle and angular velocity using accelerometers and rate gyroscopes," *Medical and Biological Engineering and Computing*, vol. 39, no. 3, pp. 294–302, 2001. [Online]. Available: http://dx.doi.org/10.1007/BF02345283

[6] M. El-Gohary, "Joint angle tracking with inertial sensors," Ph.D. dissertation, Portland State University, 2013.

[7] T. Seel, J. Raisch, and T. Schauer, "Imu-based joint angle measurement for gait analysis," *Sensors*, vol. 14, no. 4, p. 6891, 2014. [Online]. Available: http://www.mdpi.com/1424-8220/14/4/6891

[8] P.-C. Lin, H. Komsuoglu, and D. Koditschek, "Sensor data fusion for body state estimation in a hexapod robot with dynamical gaits," *Robotics, IEEE Transactions on*, vol. 22, no. 5, pp. 932–943, Oct 2006.

[9] P.-C. Lin, J.-C. Lu, C.-H. Tsai, and C.-W. Ho, "Design and implementation of a nine-axis inertial measurement unit," *Mechatronics, IEEE/ASME Transactions on*, vol. 17, no. 4, pp. 657–668, Aug 2012.

[10] P. Cheng and B. Oelmann, "Joint-angle measurement using accelerometers and gyroscopesa survey," *Instrumentation and Measurement, IEEE Transactions on*, vol. 59, no. 2, Oct 2010.

[11] G. Santaera, E. Luberto, A. Serio, M. Gabiccini, and A. Bicchi, "Low-cost, fast and accurate reconstruction of robotic and human postures via imu measurements," in *Robotics and Automation (ICRA), 2015 IEEE International Conference on*, May 2015, pp. 2728–2735.

[12] J. Vihonen, J. Honkakorpi, J. Mattila, and A. Visa, "Geometry-aided mems motion state estimation for multi-body manipulators," in *Advanced Intelligent Mechatronics (AIM), 2013 IEEE/ASME International Conference on*, July 2013, pp. 341–347.

[13] ——, "Geometry-aided angular acceleration sensing of rigid multi-body manipulator using mems rate gyros and linear accelerometers," in *Intelligent Robots and Systems (IROS), 2013 IEEE/RSJ International Conference on*, Nov 2013, pp. 2514–2520.

[14] J. Vihonen, J. Honkakorpi, J. Koivumaki, J. Mattila, and A. Visa, "Geometry-aided low-noise angular velocity sensing of rigid-body manipulator using mems rate gyros and linear accelerometers," in *Advanced Intelligent Mechatronics (AIM), 2014 IEEE/ASME International Conference on*, July 2014, pp. 570–575.

[15] J. Honkakorpi, J. Vihonen, and J. Mattila, "Mems-based state feedback control of multi-body hydraulic manipulator," in *Intelligent Robots and Systems (IROS), 2013 IEEE/RSJ International Conference on*, Nov 2013, pp. 4419–4425.

[16] X. Xinjilefu, "State estimation for humanoid robots," Ph.D. dissertation, Robotics Institute, Carnegie Mellon University, 2015.

[17] X. Xinjilefu, S. Feng, and C. Atkeson, "A distributed mems gyro network for joint velocity estimation," in *Robotics and Automation (ICRA), 2016 IEEE International Conference on (Accepted)*, 2015.

[18] W. Kabsch, "A solution for the best rotation to relate two sets of vectors," *Acta Crystallographica Section A*, vol. 32, no. 5, pp. 922–923, Sep 1976. [Online]. Available: http://dx.doi.org/10.1107/S0567739476001873